\documentclass[runningheads]{llncs}

\usepackage[T1]{fontenc}
\usepackage{fmtcount}
\usepackage{graphicx,verbatim}
\usepackage{amssymb}
\usepackage{amsmath}
\usepackage{amssymb}
\usepackage{multirow}
\usepackage{tabularx}
\usepackage{hyperref}
\usepackage{url}
\usepackage[table,xcdraw]{xcolor}
\usepackage{amsmath}
\usepackage{xcolor}
\usepackage{graphicx} 
\usepackage{booktabs}
\usepackage[misc]{ifsym}
\usepackage{color}

\begin{document}
\title{ADAptation: Reconstruction-based Unsupervised Active Learning for Breast Ultrasound Diagnosis}

\author{Yaofei Duan\inst{1} \and
Yuhao Huang \inst{2} \and
Xin Yang \inst{2} \and
Luyi Han \inst{3} \and
Xinyu Xie \inst{1} \and 
Zhiyuan Zhu \inst{2} \and
Ping He \inst{4} \and
Ka-Hou Chan \inst{1} \and
Ligang Cui \inst{4} \and
Sio-Kei Im \inst{1} \and
Dong Ni \inst{2} \and \\
Tao Tan \inst{1}\textsuperscript{(\Letter)}
}
%D
\authorrunning{Y. Duan et al.}
\institute{Faculty of Applied Sciences, Macao Polytechnic University, Macau, China\\
\email{taotan@mpu.edu.mo}\\ \and
Medical Ultrasound Image Computing (MUSIC) Lab, School of Biomedical Engineering, Medical School, Shenzhen University, Shenzhen, China
\and
Netherlands Cancer Institute, Amsterdam, Netherlands \and
Department of Ultrasound, Peking University Third Hospital, Beijing, China
}

\titlerunning{ADAptation}

\maketitle           % typeset the header of the contribution

\begin{abstract}
Deep learning-based diagnostic models often suffer performance drops due to distribution shifts between training (source) and test (target) domains. 
Collecting and labeling sufficient target domain data for model retraining represents an optimal solution, yet is limited by time and scarce resources. 
Active learning (AL) offers an efficient approach to reduce annotation costs while maintaining performance, but struggles to handle the challenge posed by distribution variations across different datasets. 
In this study, we propose a novel unsupervised \textbf{A}ctive learning framework for \textbf{D}omain \textbf{A}da\textbf{ptation}, named \textbf{ADAptation}, which efficiently selects informative samples from multi-domain data pools under limited annotation budget. 
As a fundamental step, our method first utilizes the distribution homogenization capabilities of diffusion models to bridge cross-dataset gaps by translating target images into source-domain style. 
We then introduce two key innovations: (a) a hypersphere-constrained contrastive learning network for compact feature clustering, and (b) a dual-scoring mechanism that quantifies and balances sample uncertainty and representativeness. 
Extensive experiments on four breast ultrasound datasets (three public and one in-house/multi-center) across five common deep classifiers demonstrate that our method surpasses existing strong AL-based competitors, validating its effectiveness and generalization for clinical domain adaptation. The code is available at the anonymized link: \url{https://github.com/miccai25-966/ADAptation}.

\end{abstract}

\begin{table}[t]
\centering
\caption{Quantitative analysis of cross-domain similarity score distributions Pre- and Post-reconstruction distance homogenization in breast ultrasound datasets.}
\resizebox{0.88\textwidth}{!}{
\begin{tabular}{l|cc|cc|cc|c}
\hline
                         & \multicolumn{2}{c|}{Amount} & \multicolumn{2}{c|}{Original distribution}                               & \multicolumn{2}{c|}{Homogenized distribution}    &                                                                               \\ \cline{2-7}
\multirow{-2}{*}{Domain} & \multicolumn{1}{c|}{Benign}      & Maligant      & \multicolumn{1}{c|}{Mean+Std}                              & Domain Bias & \multicolumn{1}{c|}{Mean+Std}      & Domain Bias & \multirow{-2}{*}{\begin{tabular}[c]{@{}c@{}}Public \\ available\end{tabular}} \\ \hline
Source BUSI \cite{BUSI}             & \multicolumn{1}{c|}{210}         & 437           & \multicolumn{1}{c|}{\cellcolor[HTML]{EFEFEF}0.8744+0.0015} & -           & \multicolumn{1}{c|}{-}             & -           & yes                                                                           \\
Target BUS-BRA \cite{busbra}           & \multicolumn{1}{c|}{1,268}       & 607           & \multicolumn{1}{c|}{0.8048+0.0017}                         & 0.0696      & \multicolumn{1}{c|}{0.8409+0.0014} & 0.0335      & yes                                                                           \\
Target UDIAT \cite{UDIAT}            & \multicolumn{1}{c|}{110}         & 53            & \multicolumn{1}{c|}{0.8554+0.0012}                         & 0.0190      & \multicolumn{1}{c|}{0.8609+0.0011} & 0.0135      & yes                                                                           \\
Target MC-BUS            & \multicolumn{1}{c|}{272}         & 116           & \multicolumn{1}{c|}{0.9312+0.0007}                         & 0.0568      & \multicolumn{1}{c|}{0.9091+0.0008} & 0.0347      & no                                                                            \\ \hline
\end{tabular}
}
\label{t0}
\end{table}

\begin{figure}[t]
\centering
\includegraphics[width=1\textwidth]{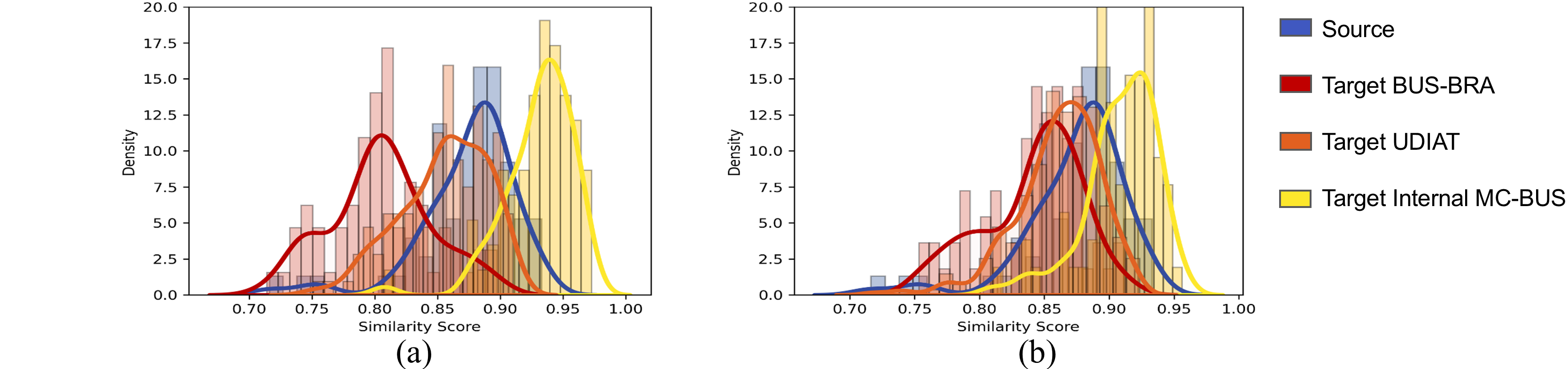}
\caption{Kernel Density Estimation (KDE) of the source domain and three target domains. (a) Original data distributions. (b) Homogenized distributions after diffusion model-based reconstruction.} 
\label{fig0}
\end{figure}

\section{Introduction}
Deep learning (DL) has revolutionized medical image analysis, yet models trained on source domains often struggle to generalize to target domains due to domain shift~\cite{huang2022online,lin}. 
This challenge is particularly pronounced in clinical settings, where variations in imaging equipments, scanning protocols, and patient populations across healthcare institutions significantly impact model performance~\cite{huang2023fourier,zhang2024unimrisegnet}. 

While supervised domain adaptation (SDA) offers solutions through transfer learning and fine-tuning, it remains impractical given the substantial time and expertise required for data annotation~\cite{guan2021domain}. 
Unsupervised domain adaptation (UDA) methods have emerged to learn domain-invariant features without target domain labels~\cite{feng2023unsupervised,uda}. 
However, existing UDA methods often lack effective sample selection mechanisms, potentially missing crucial informative samples for enhanced adaptation performance. 
These limitations highlight a fundamental trade-off in medical domain adaptation (DA) between annotation costs and model adaptation, raising a critical question: \textit{How can we optimize sample selection to maximize adaptation effectiveness with minimal annotation effort?}

Active learning (AL) emerges as a promising paradigm to address this issue by intelligently selecting the most informative samples for annotation. While traditional AL approaches focus on either representativeness \cite{jin2023label,linmans2023predictive} or uncertainty-based \cite{karamcheti2021mind} sampling strategies. The former faces annotation redundancy, while the latter may introduce distribution misalignment. Moreover, they typically assume shared feature distributions across domains, neglecting the critical DA problems in medical imaging. Recent work~\cite{ash2019deep} has begun to bridge AL with DA, inspiring subsequent research to decompose image features into domain-specific and task-specific components for unsupervised AL (UAL)~\cite{mahapatra2024alfredo}. However, the decoupling-driven solution lacks explicit modeling between source and target domains, resulting in poor interpretability and generalization.

To address these issues, we proposed ADAptation, a novel framework for unsupervised sample selection across multiple target domains. Our approach is motivated by a key insight: while source and target data exhibit distinct distributional characteristics, diffusion models~\cite{dm} can minimize domain-specific variations through reconstruction. Building upon this observation (Table \ref{t0}, Fig. \ref{fig0}), our ADAptation framework makes three key contributions: First, we integrated reconstruction-based prior knowledge in contrastive learning (CL) with hypersphere constraints for robust label-free representation. Second, we proposed a dual-scoring selection strategy to address the trade-off between sample uncertainty and representativeness. Last, we validated ADAptation on large-scale breast ultrasound (US) images from three public and one in-house multi-center datasets, efficiently handling clinical DA tasks across five DL models.

%-----fig
\begin{figure}[t]
\centering
\includegraphics[width=1\textwidth]{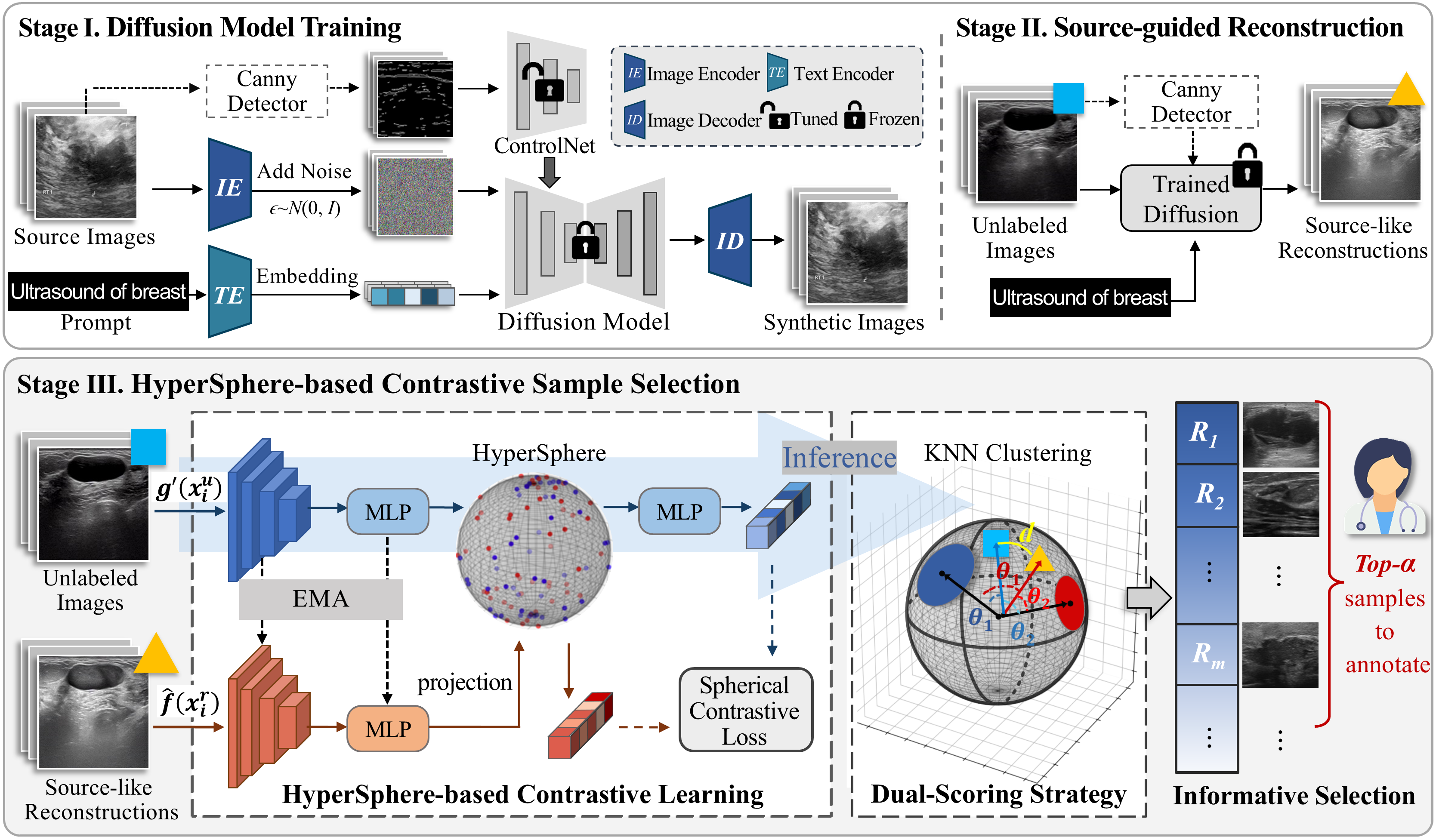}
\caption{Overview of the ADAptation framework for informative samples selection.
} 
\label{fig1}
\end{figure}

\section{Method}
We propose to integrate UAL with DA to improve breast US image classification across multiple domains, and the selected samples are generalized to fine-tune the diverse diagnostic models. 
Given a labeled source dataset $D_{S}=\left\{x_{i}^{s}, y_{i}^{s}\right\}_{i=1}^{N_{s}}$ and multiple unlabeled target domain datasets $D_{T}=\left\{x_{i}^{t}\right\}_{i=1}^{N_{t}}$, ADAptation aims to select the top $\alpha$\% most informative samples from the unlabeled data pool $D_{U} = \{D_{T1}\cup D_{T2}\cup.. \}$ for expert annotation. 
As illustrated in Fig. \ref{fig1}, ADAptation includes three stages. 
In stage I, we fine-tune a diffusion model with ControlNet~\cite{control} on source domain data. 
During Stage II, the frozen diffusion model generates source-like reconstructions for the unlabeled target images. 
In Stage III, we introduce an unsupervised CL to embed the target data and reconstructions within a normalized hypersphere.
Last, a sphere-based rule quantifies informativeness to select Top-$\alpha\%$ samples for annotation. 
It is highlighted that unlike traditional AL methods, which rely on multiple rounds of incremental learning for a single model, our method achieves effective single-iteration approach across multiple models. 
This design addresses clinical needs where diagnostic models require rapid updates with new data. Our ADAptation framework provides a more generalizable solution for efficient model adaptation in clinical settings.

\subsection{Source-guided Reconstruction for Domain Alignment} 
Due to serious domain gaps between source and target data, most previous AL methods are insufficient and potentially biased in identifying informative samples.
To bridge this domain gap, we propose a source-guided reconstruction strategy based on diffusion models~\cite{dm} in Stage I and II. Our key insight is that by conditioning the generation process on both source domain knowledge and target domain structural canny edge map priors, we can synthesize source-like reconstructions while preserving critical medical characteristics of target US images. 
Formally, the reconstruction process is formulated as:
\begin{equation}
\begin{aligned}
    \mathcal{L} & =\mathbb{E}_{\boldsymbol{z}_{0}^S, \boldsymbol{t}, \boldsymbol{c}_{p}^S, \boldsymbol{c}_{f}^S, \epsilon \sim \mathcal{N}(0,1)}\left[\| \epsilon-\epsilon_{\theta}\left(\boldsymbol{z}_{t}^S, \boldsymbol{t}, \boldsymbol{c}_{p}^S, \boldsymbol{c}_{\mathrm{f}}^S\right) \|_{2}^{2}\right], & \text{(training)} \\
    \boldsymbol{z}_{0}^T & = \text{Sampling}\left(\epsilon_{\theta}\left(\boldsymbol{z}_{t}^T, \boldsymbol{t}, \boldsymbol{c}_{p}^T, \boldsymbol{c}_{\mathrm{f}}^T\right)\right), \quad \boldsymbol{z}_{t}^T \sim \mathcal{N}(0,1), & \text{(inference)}
\end{aligned}
\end{equation}
where $S$ and $T$ denote Source and Target, ${z}_{0}$ is input image, $t$ is timestep, ${c}_{p}$ and ${c}_{f}$ are the prompt and canny edge, respectively. 
We use prompt "Ultrasound of breast" as a prior semantic anchor to align image features with relevant medical concepts, and further replace the original text encoder (i.e., CLIP) with BiomedCLIP \cite{biomedclip} for better semantic alignment. Finally, the reconstructions effectively approximate the source distribution while maintaining self-characteristics, enabling unbiased AL selection.

\subsection{HyperSphere Representation for Contrastive Learning}
CL has proven effective in capturing high-level representations on unsupervised tasks~\cite{Zhang_2024_WACV}. 
Inspired by \cite{byol}, we incorporate a teacher network $\hat{f}\left(x_{i}^{u}\right)$ for and student $g^{\prime}\left(x_{i}^{r}\right)$ network in Stage III to minimize feature discrepancies between US images and their reconstructions. 
This alignment encourages the network to learn robust feature independent of any specific domain.
We employ a ResNet-50 backbone pre-trained on source data for initial feature extraction, followed by MLP to enhance representation capacity.
However, in cross-domain scenarios, direct feature learning often leads to scattered representations due to domain shifts. 
Therefore, we introduce hypersphere constraint by projecting embeddings $\mathbf{z}$ onto a 255-dimensional hypersphere $\mathbf{z} \in \mathbb{R}^{256}$ through L2 norm:
% To ensure compact representations, we project the embeddings $\mathbf{z}$ onto a 255-dimensional hypersphere $\mathbf{z} \in \mathbb{R}^{256}$ through L2 norm: 
\begin{equation}
\hat{f}\left(x_{i}^{u}\right)=\frac{f\left(x_{i}^{u}\right)}{\left\|f\left(x_{i}^{u}\right)\right\|_{2}}, \qquad
g^{\prime}\left(x_{i}^{r}\right)=\frac{g\left(x_{i}^{r}\right)}{\left\|g\left(x_{i}^{r}\right)\right\|_{2}}.
\end{equation}

This can map cross-domain features onto a fixed-length manifold, preventing domain-specific bias and promoting unsupervised discriminative feature learning.

To optimize this geometry-aware representation, we introduce a spherical contrastive loss with two key components: (1) Angular Contrastive Loss minimizes angular discrepancies between teacher and student network representations. (2) Angular Scaling Factor adjusts the penalty on angular differences to balance alignment precision and generalization. The total loss is defined as:
\begin{equation}
    L(\hat{f}\left(x\right), g^{\prime}\left(x\right))=\frac{1}{N} \sum_{i=1}^{N}\left(m \cdot \arccos \left(\frac{f\left(x_{i}^{u}\right) \cdot g\left(x_{i}^{r}\right)}{\left\|f\left(x_{i}^{u}\right)\right\| \cdot \left\|g\left(x_{i}^{r}\right)\right\|}\right)\right)^{2},
\end{equation}
where $N$ denotes the batch size, and $m=4$ represents the adaptive scaling factor. During inference, only the frozen-weight student network is employed.

\subsection{Informative Sample Selection via Dual-Scoring}
A balanced consideration of uncertainty and representativeness in AL sample selection is crucial for improving the diagnostic performance of diverse downstream models. 
We propose a dual-scoring strategy that considers both two sides to select the most informative samples from the unlabeled target pool. 

On one hand, we employ KNN clustering with $k$ centroids in the hyperspherical feature space to estimate uncertainty. Given the angular differences $\{\theta_{1},\theta_{2},...,\theta_{k}\}$ between unlabeled image $x_{i}^{u}$ and all centroids, the uncertainty score is computed as the absolute value between the smallest and largest angular differences, where larger difference indicates the data point is closer to a specific centroid, while smaller value indicates higher uncertainty. 
On the other hand, the representativeness score measures the divergence from the source distribution via the spherical distance between the unlabeled sample $x_{i}^{u}$ and its reconstruction $x_{i}^{r}$. Subsequently, we formulate the Informative score $I_{i}$ as a weighted combination of the two aforementioned metric ranks, which can be formulated as:
\begin{equation}
    I_{i} = \arg min _{p, q \in\{1, \ldots, k\}}\left|\theta_{p}-\theta_{q}\right| +  \omega  \times \text { SphericalDist}\left(x_{i}^{u}, x_{i}^{r}\right),
\end{equation}
where $\text {SphericalDist}$ equals $\text {arccos}(\cdot)$.
Finally, the target samples are ranked in ascending order of $I_{i}$, with top-ranked candidates selected for expert annotation:

\begin{equation}
   \mathcal{S}=\operatorname{top}-\alpha \%\left(\left\{x_{i}^{u} \mid I_{i}\right\}_{i=1}^{N}\right),
\end{equation}
where $S$ denotes the set of selected samples, $\alpha \%\in(0,1)$ represents the selection ratio, $N$ is the total number of unlabeled samples from diverse domains.

\section{Experiments}
\textbf{Dataset and Implementation details.}
We evaluated the ADAptation framework on three public breast US datasets and one internal multi-center dataset (MC-BUS), details refer to Table~\ref{t0}. Specifically, the BUSI dataset served as the source domain, with 90\% samples used for training and 10\% reserved for extra reconstruction qualitative analysis. 
To simulate multi-domain AL scenarios, we utilized UDIAT, BUS-BRA, and MC-BUS as target domains, with each split into selection (90\%) and test (10\%) sets. 
The selection sets formed the unlabeled target pools for AL, while the test sets were used for performance evaluation.

All models were implemented in \textit{PyTorch} and trained on NVIDIA A40 GPU with 48GB memory. 
Data augmentation includes horizontal flipping and rotation. 
Our framework was trained for 200 epochs in both Stages I\&III with a learning rate (\textit{lr}) of 1e-4. 
Then, the downstream classifiers were initialized with source pre-trained weights and fine-tuned on selected data for 140 epochs.
Adam optimizer was used, with a 0.001 initial \textit{lr} and a batch size of 8. 
% To ensure efficient learning, 
Besides, the cosine annealing schedule was leveraged to adjust \textit{lr} dynamically.

\begin{table}[t] 
\centering 
\caption{Comparison of classification accuracy between ADAptation and other AL methods on target domain test sets. (\textbf{Bold} represents the best result, \underline{Underline} represents the second best result).}
\renewcommand{\arraystretch}{1} %
\resizebox{\textwidth}{!}{
\begin{tabular}{l|l|ccccc|c}
% \hline
\hline
\textbf{\begin{tabular}[c]{@{}l@{}}Annotation \\ Ratio\end{tabular}} & \textbf{AL methods} & \textbf{ResNet-50}~\cite{resnet}                     & \textbf{DenseNet-169} \cite{dense}                  & \textbf{ShuffleNet} \cite{shuff}                   & \textbf{MobileNet} \cite{mobile}                    & \textbf{EfficientNet} \cite{eff}                 & \textbf{Average}               \\ \hline
\cellcolor[HTML]{D9E1F2}100\%                                                                & \cellcolor[HTML]{D9E1F2}Full                & \cellcolor[HTML]{D9E1F2}0.9229{\footnotesize±0.0169} & \cellcolor[HTML]{D9E1F2}0.9193{\footnotesize±0.0169} & \cellcolor[HTML]{D9E1F2}0.9668{\footnotesize±0.0170} & \cellcolor[HTML]{D9E1F2}0.9488{\footnotesize±0.0103} & \cellcolor[HTML]{D9E1F2}0.9595{\footnotesize±0.0314} & \cellcolor[HTML]{D9E1F2}0.9435 \\ \hline
     & \multicolumn{1}{l|}{\cellcolor[HTML]{FFF7DD}Random}              & \cellcolor[HTML]{FFF7DD}0.6555{\footnotesize±0.0476}                         & \cellcolor[HTML]{FFF7DD}0.6628{\footnotesize±0.0494}                         & \cellcolor[HTML]{FFF7DD}0.7725{\footnotesize±0.0403}                         & \cellcolor[HTML]{FFF7DD}0.7540{\footnotesize±0.0421}                         & \cellcolor[HTML]{FFF7DD}0.7653{\footnotesize±0.0403}                         & \cellcolor[HTML]{FFF7DD}0.7220                         \\
         & VAAL {\cite{vaal}}                 & 0.6561{\footnotesize±0.0476}                         & 0.6889{\footnotesize±0.0476}                         & 0.7540{\footnotesize±0.0439}                         & 0.7547{\footnotesize±0.0439}                         & 0.7684{\footnotesize±0.0421}                         & 0.7281                         \\
         & Core-Set {\cite{core}}            & 0.7037{\footnotesize±0.0476}                         & 0.7280{\footnotesize±0.0421}                         & 0.7830{\footnotesize±0.0403}                         & 0.7586{\footnotesize±0.0439}                         & 0.7950{\footnotesize±0.0403}                         & 0.7537                         \\
     & Max-Entropy {\cite{entro}}         & 0.7070{\footnotesize±0.0476}                         & 0.6995{\footnotesize±0.0458}                         & \underline{0.8282{\footnotesize±0.0366}}                         & 0.7657{\footnotesize±0.0421}                         & 0.7984{\footnotesize±0.0403}                         & 0.7598                         \\
         & BALD {\cite{bald}}                & 0.7033{\footnotesize±0.0441}                         & \underline{0.7367{\footnotesize±0.0439}}                         & 0.8133{\footnotesize±0.0403}                         & 0.7661{\footnotesize±0.0403}                         & \underline{0.8061{\footnotesize±0.0403}}                         & 0.7651                         \\
     & LfOSA {\cite{last}}                  & \underline{0.7437{\footnotesize±0.0376}}                         & 0.7347{\footnotesize±0.0421}                         & 0.7973{\footnotesize±0.0342}                         & \underline{0.7704{\footnotesize±0.0403}}                         & \multicolumn{1}{c|}{0.7897{\footnotesize±0.0405}}                         & \underline{0.7672}                      \\
\multirow{-7}{*}{20\%}                                               & \textbf{ADAptation} & \textbf{0.7816{\footnotesize±0.0403}}                & \textbf{0.8023{\footnotesize±0.0403}}                & \textbf{0.8343{\footnotesize±0.0382}}                & \textbf{0.8090{\footnotesize±0.0430}}                & \textbf{0.8135{\footnotesize±0.0348}}                & \textbf{0.8081}                \\ \hline
     & \multicolumn{1}{l|}{\cellcolor[HTML]{FFF7DD}Random}              & \cellcolor[HTML]{FFF7DD}0.6831{\footnotesize±0.0324}                         & \cellcolor[HTML]{FFF7DD}0.7357{\footnotesize±0.0415}                         & \cellcolor[HTML]{FFF7DD}0.8313{\footnotesize±0.0387}                         & \cellcolor[HTML]{FFF7DD}0.8064{\footnotesize±0.0487}                         & \cellcolor[HTML]{FFF7DD}0.8200{\footnotesize±0.0396}                         & \cellcolor[HTML]{FFF7DD}0.7753                         \\
     & VAAL {\cite{vaal}}                & 0.6887{\footnotesize±0.0370}                         & 0.7362{\footnotesize±0.0407}                         & 0.8019{\footnotesize±0.0431}                         & 0.8087{\footnotesize±0.0372}                         & 0.7995{\footnotesize±0.0437}                         & 0.7670                         \\
     & Core-Set {\cite{core}}            & 0.7196{\footnotesize±0.0424}                         & 0.7347{\footnotesize±0.0351}                         & 0.8423{\footnotesize±0.0494}                         & 0.8089{\footnotesize±0.0370}                        & 0.8307{\footnotesize±0.0412}                         & 0.7872                         \\
     & Max-Entropy {\cite{entro}}         & 0.7211{\footnotesize±0.0320}                         & 0.7357{\footnotesize±0.0415}                         & \underline{0.8567{\footnotesize±0.0289}}                         & \underline{0.8315{\footnotesize±0.0487}}                         & 0.8205{\footnotesize±0.0377}                         & \underline{0.7931}                         \\
     & BALD {\cite{bald}}                & 0.7255{\footnotesize±0.0287}                         & \underline{0.7524{\footnotesize±0.0300}}                         & 0.8315{\footnotesize±0.0374}                         & 0.8017{\footnotesize±0.0370}                         & \underline{0.8271{\footnotesize±0.0280}}                         & 0.7876                         \\
     & LfOSA {\cite{last}}                  & \underline{0.7607{\footnotesize±0.0403}}                         & 0.7492{\footnotesize±0.0387}                         & 0.8361{\footnotesize±0.0403}                         & 0.7859{\footnotesize±0.0342}                         & \multicolumn{1}{c|}{0.8092{\footnotesize±0.0403}}                         & 0.7882                               \\
\multirow{-7}{*}{30\%}                                               & \textbf{ADAptation} & \textbf{0.7966{\footnotesize±0.0314}}                & \textbf{0.8085{\footnotesize±0.0306}}                & \textbf{0.8429{\footnotesize±0.0487}}                & \textbf{0.8272{\footnotesize±0.0196}}                & \textbf{0.8315{\footnotesize±0.0186}}                & \textbf{0.8213}                \\ \hline
     & \multicolumn{1}{l|}{\cellcolor[HTML]{FFF7DD}Random}              & \cellcolor[HTML]{FFF7DD}0.8364{\footnotesize±0.0192}                         & \cellcolor[HTML]{FFF7DD}0.8081{\footnotesize±0.0421}                         & \cellcolor[HTML]{FFF7DD}0.8898{\footnotesize±0.0230}                         & \cellcolor[HTML]{FFF7DD}0.8717{\footnotesize±0.0351}                         & \cellcolor[HTML]{FFF7DD}0.8866{\footnotesize±0.0342}                         & \cellcolor[HTML]{FFF7DD}0.8585                         \\
     & VAAL {\cite{vaal}}                & 0.8439{\footnotesize±0.0431}                         & 0.8191{\footnotesize±0.0407}                         & 0.8759{\footnotesize±0.0407}                         & 0.8747{\footnotesize±0.0476}                         & 0.8973{\footnotesize±0.0403}                         & 0.8622                         \\
     & Core-Set {\cite{core}}            & 0.8483{\footnotesize±0.0203}                         & 0.8227{\footnotesize±0.0403}                         & 0.8947{\footnotesize±0.0396}                         & 0.8889{\footnotesize±0.0403}                        & 0.8960{\footnotesize±0.0412}                         & 0.8701                         \\
     & Max-Entropy {\cite{entro}}         & \textbf{0.8596{\footnotesize±0.0420}}                         & \underline{0.8363{\footnotesize±0.0380}}                         & 0.8742{\footnotesize±0.0320}                         & 0.8776{\footnotesize±0.0403}                         & 0.8969{\footnotesize±0.0420}                         & 0.8689                         \\
     & BALD {\cite{bald}}                & 0.8395{\footnotesize±0.0370}                         & 0.8360{\footnotesize±0.0421}                         & 0.8806{\footnotesize±0.0403}                         & 0.8856{\footnotesize±0.0370}                         & 0.8991{\footnotesize±0.0376}                         & 0.8682                         \\
     & LfOSA {\cite{last}}                  & 0.8497{\footnotesize±0.0343}                         & 0.8208{\footnotesize±0.0476}                         & \underline{0.9041{\footnotesize±0.0403}}                         & \underline{0.8917{\footnotesize±0.0312}}                         & \multicolumn{1}{c|}{\textbf{0.9056{\footnotesize±0.0403}}}                         & \underline{0.8744}                               \\
\multirow{-7}{*}{50\%}                                               & \textbf{ADAptation} & \underline{0.8566{\footnotesize±0.0241}}                & \textbf{0.8500{\footnotesize±0.0403}}                & \textbf{0.9051{\footnotesize±0.0403}}                & \textbf{0.8943{\footnotesize±0.0304}}                & \underline{0.9008{\footnotesize±0.0318}}                & \textbf{0.8814}                \\ \hline
     & \multicolumn{1}{l|}{\cellcolor[HTML]{FFF7DD}Random}              & \cellcolor[HTML]{FFF7DD}0.8981{\footnotesize±0.0203}                         & \cellcolor[HTML]{FFF7DD}\underline{0.8916{\footnotesize±0.0280}}                         & \cellcolor[HTML]{FFF7DD}\textbf{0.9415{\footnotesize±0.0234}}                         & \cellcolor[HTML]{FFF7DD}\underline{0.9313{\footnotesize±0.0365}}                         & \cellcolor[HTML]{FFF7DD}\underline{0.9536{\footnotesize±0.0403}}                         & \cellcolor[HTML]{FFF7DD}\underline{0.9232}                         \\
     & VAAL {\cite{vaal}}                & 0.8690{\footnotesize±0.0403}                         & 0.8828{\footnotesize±0.0396}                         & 0.9215{\footnotesize±0.0370}                         & 0.9289{\footnotesize±0.0298}                         & 0.9427{\footnotesize±0.0321}                         & 0.9090                         \\
     & Core-Set {\cite{core}}            & 0.8875{\footnotesize±0.0320}                         & 0.8830{\footnotesize±0.0403}                         & 0.9308{\footnotesize±0.0297}                         & 0.9265{\footnotesize±0.0403}                        & 0.9336{\footnotesize±0.0403}                         & 0.9158                         \\
     & Max-Entropy {\cite{entro}}         & 0.8940{\footnotesize±0.0424}                         & 0.8936{\footnotesize±0.0287}                         & 0.9274{\footnotesize±0.0289}                         & 0.9287{\footnotesize±0.0320}                         & 0.9351{\footnotesize±0.0403}                         & 0.9158                         \\
     & BALD {\cite{bald}}                & 0.8893{\footnotesize±0.0376}                         & 0.8861{\footnotesize±0.0342}                         & 0.9241{\footnotesize±0.0376}                         & 0.9230{\footnotesize±0.0315}                         & 0.9337{\footnotesize±0.0314}                         & 0.9112                         \\
     & LfOSA {\cite{last}}                  & \underline{0.9075{\footnotesize±0.0218}}                         & 0.8800{\footnotesize±0.0320}                         & 0.9336{\footnotesize±0.0372}                         & 0.9294{\footnotesize±0.0320}                         & \multicolumn{1}{c|}{0.9433{\footnotesize±0.0403}}                         & 0.9188                               \\
\multirow{-7}{*}{80\%}                                               & \textbf{ADAptation} & \textbf{0.9304{\footnotesize±0.0344}}                & \textbf{0.9049{\footnotesize±0.0314}}                & \underline{0.9352{\footnotesize±0.0278}}                & \textbf{0.9490{\footnotesize±0.0327}}                & \textbf{0.9558{\footnotesize±0.0314}}                & \textbf{0.9351}                \\ \hline
% \hline
\end{tabular}
}
\label{c1}
\end{table}

\textbf{Method Comparison.}
We evaluate ADAptation against state-of-the-art AL methods including both uncertainty (Max-Entropy \cite{entro}, BALD \cite{bald}, LfOSA \cite{last}) and representative-based sampling (Core-Set \cite{core}, VAAL \cite{vaal}). 
To ensure a comprehensive evaluation, we conduct experiments with varying annotation budgets 
(20\%, 30\%, 50\%, 80\%) on the binary classification task (benign or malignant). 
The selected samples are then used to fine-tune five DL models. 
The averaged classification results across target sets are reported in Table \ref{c1}.

In the low-resource scenario (20\% annotation), ADAptation achieves an average accuracy of 0.8081, significantly surpassing all competitors ($p<0.01$), and with a 4.83\% improvement over the second-best LfOSA method.  
As the annotation ratio increases to 30\% and 50\%, ADAptation maintains its performance advantage with average improvements of 3.95\% and 2.87\% respectively over the second-best methods, demonstrating its effectiveness in handling complex cross-domain scenarios, and robustness for diverse model architectures. 
The relatively narrow performance gap under the 50\% annotation setting can be attributed to the increased likelihood of selecting informative samples as the labeled data volume grows. 
Notably, in high-resource scenarios (80\%), other methods fail to achieve comprehensive coverage across heterogeneous multi-domain data pools when compared to random sampling due to biased selection strategies. 
In contrast, ADAptation's dual-score strategy lead to accuracy of 0.9351, approaching the upper bound (0.9435). 
This validates the robustness of our approach in handling domain shifts irrespective of the annotation budget and diagnostic models.

\textbf{Ablation Study.}
As shown in Table \ref{a1}, incorporating CL markedly enhances performance across all downstream models (increase 4.94\%) by improving the feature representations under unsupervised settings. The addition of hypersphere regularization further boosts performance by constraining feature embeddings in a compact latent space, facilitating better domain adaptation. Notably, the removal of reconstruction prior leads to a 0.61\% decrease, indicating its effectiveness in mitigating cross-domain discrepancies, especially in cases with large variations between source and target domains. 
We also analyze the impact of cluster numbers on ADAptation's performance. Results indicate that four clusters yield optimal performance (0.8081), which we adopt for all subsequent experiments.

\begin{table}[t]
\centering
\caption{Accuracy metric of ablation results for model components and number of cluster centers under 20\% annotation ratio, test on the target domain sets.}
\renewcommand{\arraystretch}{1.0} %
\resizebox{\textwidth}{!}{
\begin{tabular}{l|ccccc|c}
\hline
\textbf{Method}   & \textbf{ResNet-50} \cite{resnet}     & \textbf{DenseNet-169} \cite{dense}  & \textbf{ShuffleNet} \cite{shuff}    & \textbf{MobileNet} \cite{mobile}     & \textbf{EfficientNet} \cite{eff}  & \textbf{Average} \\ \hline
Frozen backbone-Only & 0.6686±0.0206           & 0.6684±0.0411          & 0.7508±0.0403          & 0.7513±0.0403          & 0.7799±0.0419          & 0.7238           \\
+CL                 & 0.7380±0.0312           & 0.7572±0.0206          & 0.8029±0.0236          & 0.7703±0.0410          & 0.7977±0.0141          & 0.7732           \\
+CL+Hypersphere      & 0.7630±0.0221           & 0.7859±0.0285          & 0.8282±0.0276          & 0.7771±0.0290          & 0.8032±0.0212          & 0.7915           \\
w/o Reconstruction   & 0.7794±0.0236           & 0.7948±0.0206          & 0.8223±0.0470          & 0.7908±0.0406          & 0.8060±0.0196          & 0.7987           \\
ADAptation (ours)           & \textbf{0.7816 ±0.0403} & \textbf{0.8023±0.0403} & \textbf{0.8343±0.0382} & \textbf{0.8090±0.0403} & \textbf{0.8135±0.0348} & \textbf{0.8081}  \\ 
\hline
\textbf{Cluster Number}   & \textbf{ResNet-50} \cite{resnet}     & \textbf{DenseNet-169} \cite{dense}  & \textbf{ShuffleNet} \cite{shuff}    & \textbf{MobileNet} \cite{mobile}     & \textbf{EfficientNet} \cite{eff}  & \textbf{Average} \\ \hline
Cluster=2 & \underline {0.7809±0.0206}                                                 & \underline {0.7986±0.0206}                                                    & \underline {0.8234±0.0285}                                                   & 0.7921±0.0224                                                  & \textbf{0.8274±0.0199}                                            & \underline {0.8045}           \\
Cluster=3 & 0.7771±0.0290                                                 & 0.7806±0.0217                                                    & 0.8009±0.0225                                                   & 0.7835±0.0236                                                  & 0.8159±0.0410                                                     & 0.7916           \\
Cluster=4 (ours) & \textbf{0.7816±0.0403}                                        & \textbf{0.8023±0.0403}                                           & \textbf{0.8343±0.0382}                                          & \textbf{0.8090±0.0403}                                         & \underline {0.8135±0.0348}                                                     & \textbf{0.8081}  \\ 
Cluster=5 & 0.7607±0.0214                                        & 0.7921±0.0352                                           & 0.7982±0.0219                                          & \underline {0.7948±0.0219}                                         & 0.8001±0.0312                                                     & 0.7892  \\ \hline
\end{tabular}
}
\label{a1}
\end{table}

\begin{table}[t]
\centering
\caption{Quantitative results for 512x512 Breast US image reconstruction on both source and target domain datasets. Statistical significance was tested with $p<.001$.}
\renewcommand{\arraystretch}{1.0} %
\resizebox{1\textwidth}{!}{
\begin{tabular}{l|lllll}
\hline
\textbf{Domian}                     & \multicolumn{1}{c}{\textbf{PSNR \textcolor{red!65!black}{↑}}}      & \multicolumn{1}{c}{\textbf{MS-SSIM \textcolor{red!65!black}{↑}}}  & \multicolumn{1}{c}{\textbf{RMSE \textcolor{blue!65!black}{↓}}}       & \multicolumn{1}{c}{\textbf{LPIPS\_Alex \textcolor{blue!65!black}{↓}}}      & \multicolumn{1}{c}{\textbf{LPIPS\_VGG \textcolor{blue!65!black}{↓}}} \\ \hline
Source BUSI             & 13.98±3.12 \footnotesize{\textcolor{gray}{(+0.00)}} & 0.47±0.09 \footnotesize{\textcolor{gray}{(+0.00)}} & 53.57±21.79 \footnotesize{\textcolor{gray}{(+0.00)}} & 0.110±0.056 \footnotesize{\textcolor{gray}{(+0.00)}} & 0.240±0.069 \footnotesize{\textcolor{gray}{(+0.00)}} \\
Target BUS-BRA         & 11.31±2.14 \footnotesize{\textcolor{red!90!black}{(-2.67)}}          & 0.26±0.09 \footnotesize{\textcolor{red!90!black}{(-0.21)}}         & 71.08±16.58 \footnotesize{\textcolor{blue!85!black}{(-17.51)}}          & 0.178±0.077 \footnotesize{\textcolor{blue!85!black}{(-0.068)}}         & 0.299±0.052 \footnotesize{\textcolor{blue!85!black}{(-0.059)}}        \\
Target UDIAT            & 11.89±2.26 \footnotesize{\textcolor{red!90!black}{(-2.34)}}         & 0.23±0.08 \footnotesize{\textcolor{red!90!black}{(-0.24)}}         & 68.37±18.28 \footnotesize{\textcolor{blue!85!black}{(-14.80)}}         & 0.131±0.052 \footnotesize{\textcolor{blue!85!black}{(-0.021)}}         & 0.249±0.048 \footnotesize{\textcolor{blue!85!black}{(-0.009)}}         \\
Target MC-BUS & 10.40±1.63 \footnotesize{\textcolor{red!90!black}{(-3.58)}}         & 0.37±0.09 \footnotesize{\textcolor{red!90!black}{(-0.11)}}         & 77.95±15.39 \footnotesize{\textcolor{blue!85!black}{(-24.38)}}         & 0.295±0.103 \footnotesize{\textcolor{blue!85!black}{(-0.185)}}         & 0.431±0.098 \footnotesize{\textcolor{blue!85!black}{(-0.191)}}         \\

\hline
\end{tabular}
}
\label{t4}
\end{table}

\textbf{Quantitative Evaluation of Reconstruction Stage.}
We quantitatively evaluated reconstruction results on source and target domain datasets using pixel- and feature-level metrics, as shown in Table \ref{t4}. The results demonstrate significantly lower reconstruction quality on the target domain, highlighting domain bias. Furthermore, the reconstructions can serve as valuable prior knowledge for constructing CL frameworks to improve domain adaptation.

\begin{figure}[t]
\centering
\includegraphics[width=1\textwidth]{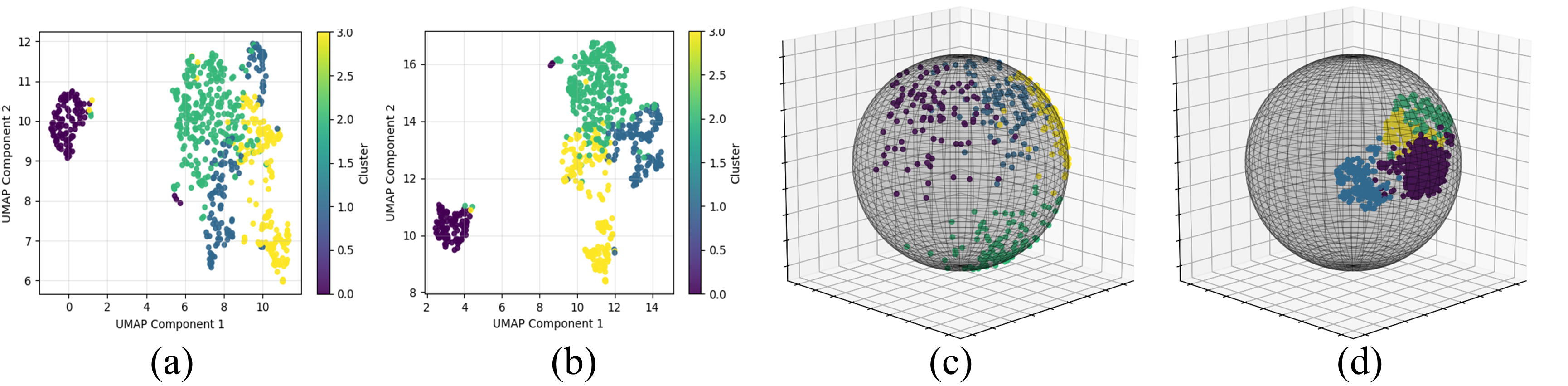}
\caption{Visualizations of Feature Projections: (a) Data embedding using KNN; (b) Embedding with Hypersphere constraint applied; (c) 3D spherical visualization result before clustering; (d) 3D spherical visualization result after clustering.} 
\label{vis1}
\end{figure}

\begin{figure}[t] 
\centering
\includegraphics[width=1\textwidth]{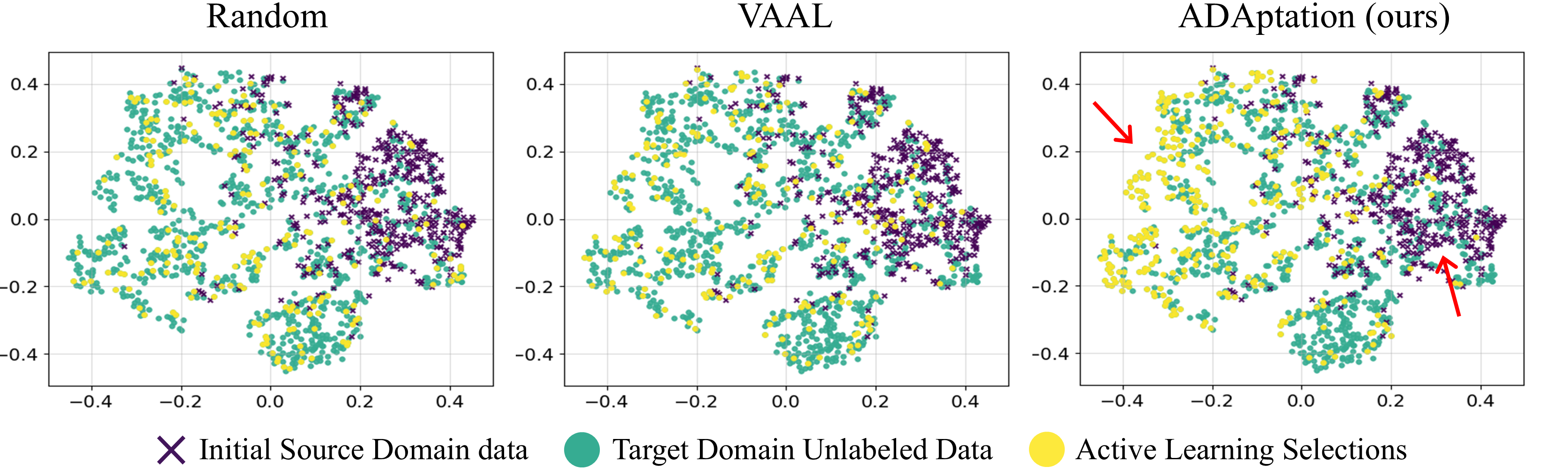}
\caption{T-SNE visualizations of different AL sampling strategies on breast US. The black, green, and yellow symbols represent the source, target, and selection data points.} 
\label{vis2}
\end{figure}

\textbf{Qualitative analysis.}
Fig. \ref{vis1} shows the initial embedding with KNN, where the clusters appear dispersed (a). 
After applying hypersphere constraints, the embeddings exhibit significantly improved compactness and separation (b). 
This is further validated by projecting the 255-D embeddings onto a 3D sphere for visualization (Fig. \ref{vis1}(c)(d)), highlighting the transformation from scattered distributions to spatially coherent and well-structured clusters. 
We further analyze the effectiveness of different sampling strategies through T-SNE visualizations (Fig. \ref{vis2}), which show the feature distributions of breast US images (taken from the fully connected layer of the trained ResNet-50). 
Random shows uniform coverage across the feature space, and VAAL is limited in boundary region coverage and tends to select samples that potentially overlap with the labeled source domain, compromising annotation efficiency. 
In contrast, ADAptation demonstrates a more strategic sample selection. 
Specifically, it effectively identifies informative samples that are well-distributed across the unlabeled manifold (\textit{Representativeness}) while emphasizing boundary samples in the target distribution (\textit{Uncertainty}), as indicated by red arrows. These boundary samples are particularly valuable as they represent challenging cases in breast US images where the diagnostic models exhibit higher misclassification rates. 
Our dual-score selection, which balances feature space coverage with uncertainty sampling, enables efficient DA while minimizing annotation costs.

\section{Conclusion}
In this study, we explore the UAL method to address the DA challenge in medical image analysis. We propose a novel framework called ADAptation, incorporating diffusion model for first turning the target images into source style, and introducing a teacher-student network for enhanced feature representation and a dual-score strategy for efficient sample selection. Extensive validation demonstrates superior classification performance improvement with limited labels, significantly reducing the annotation burden. Future work will focus on extending to different modalities and various downstream clinical tasks.

    % The following acknowledgement and disclaimer sections should be removed for the double-blind review process.  
    % If and when your paper is accepted, reinsert the acknowledgement and the disclaimer clause in your final camera-ready version.

\begin{credits}
\subsubsection{\ackname} This work was supported by the grant from Science and Technology Development Fund of Macao (0021/2022/AGJ), National Natural Science Foundation of China (12326619, 62101343, 62171290, 82201851), Science and Technology Planning Project of Guangdong Province (2023A0505020002), Frontier Technology Development Program of Jiangsu Province (BF2024078), Shenzhen-Hong Kong Joint Research Program (SGDX20201103095613036), Guangxi Province Science Program (2024AB17023), and Multi-center Clinical Study of Intelligent Prenatal Ultrasound (ChiCTR2300071300).

\subsubsection{\discintname}
The authors have no competing interests to declare that are relevant to the content of this article.
\end{credits}

\bibliographystyle{splncs04}
\bibliography{Paper-0966}

\end{document}